\documentclass[letterpaper, 10 pt, conference]{ieeeconf}  

\usepackage[T1]{fontenc}
\usepackage[table,xcdraw]{xcolor}
\usepackage{colortbl}
\usepackage[normalem]{ulem}
\usepackage{booktabs}
\useunder{\uline}{\ul}{}
\usepackage{array}
\usepackage{tabularx}
\usepackage{array}
\usepackage{cuted}
\usepackage{placeins}
\usepackage{capt-of}
\definecolor{VLABlue}{RGB}{241,246,250}
\definecolor{WAMGreen}{RGB}{241,248,243}
\definecolor{AblYellow}{RGB}{250,247,232}
\definecolor{DeepRed}{RGB}{140,45,45}
\newcolumntype{L}{>{\raggedright\arraybackslash}X}
\newcolumntype{S}{c<{\hspace{1.3pt}}}
\newcolumntype{M}{>{\hspace{1.3pt}}c<{\hspace{1.3pt}}}
\newcolumntype{A}{>{\hspace{1.3pt}}c}
\newcolumntype{Y}{>{\centering\arraybackslash}X}

\IEEEoverridecommandlockouts                              

\usepackage{graphicx} 
\usepackage{amsmath} 
\usepackage{amssymb}  

\title{\LARGE \bf
MoWAM: Explicit Future Motion Prediction for Efficient \\World Action Models
}

\author{Anonymous Authors}

\author{
Jiayu Wang$^{1}$, Bin Zhu$^{2}$, Yue Yu$^{1}$, and Jingjing Chen$^{3*}$%
\thanks{$^{*}$Corresponding author.}%
\thanks{$^{1}$Jiayu Wang and Yue Yu are with the College of Computer Science and Artificial Intelligence,
        Fudan University, Shanghai, China
        {\tt\small \{jiayuwang25,yuy24\}@m.fudan.edu.cn}}%
\thanks{$^{2}$Bin Zhu is with Singapore Management University,
        Singapore
        {\tt\small binzhu@smu.edu.sg}}%
\thanks{$^{3}$Jingjing Chen is with the Institute of Trustworthy Embodied AI,
        Fudan University, Shanghai, China
        {\tt\small chenjingjing@fudan.edu.cn}}%
}

\begin{document}

\maketitle
\thispagestyle{empty}
\pagestyle{empty}

\begin{abstract}
World Action Models (WAMs) improve robot policy learning by incorporating future dynamics, yet explicitly generating future videos at inference introduces substantial computational overhead. Removing future generation improves efficiency, but leaves future dynamics only implicitly encoded in observation features, which can limit robustness under distribution shifts. We propose MoWAM, an efficient WAM that replaces future video generation with explicit future motion prediction. Instead of reconstructing the complete future scene, MoWAM models structured robot motion as a compact abstraction of the future, capturing how the robot is expected to evolve under the current scene and interaction constraints. A Mixture-of-Transformer architecture learns future visual dynamics during training while jointly predicting motion and action, allowing video generation to be removed entirely at inference while retaining an explicit representation of the future. The compact motion representation further enables efficient inference-time scaling by sampling multiple candidates of motion and action pairs and selecting among them with a motion-aware task-progress verifier. Experiments on LIBERO, LIBERO-Plus, and real-world manipulation tasks demonstrate that MoWAM achieves strong in-distribution performance, improved out-of-distribution robustness, and higher average real-world success than representative WAM baselines. In addition, performance improves as more candidates are explored, demonstrating that explicit future motion provides an effective and efficient basis for inference-time scaling.

\end{abstract}

\section{INTRODUCTION}
Vision-Language-Action (VLA) models~\cite{brohan2023rt2, intelligence2025pi05, kim2024openvla, kim2025fine, pertsch2025fast, wang2026unified} have emerged as a promising paradigm for general-purpose robot control by mapping visual observations and language instructions directly to low-level actions. Despite their strong policy-learning capabilities, most VLAs primarily model what action to execute, without explicitly reasoning about how the scene will evolve as the action is carried out. World Action Models (WAMs) ~\cite{guo2024prediction, zhu2025unified, Bi_2026_CVPR, ye2026world, cen2025worldvla} address this limitation by incorporating future world prediction into policy learning. By leveraging the spatiotemporal priors of video generative models~\cite{wan2025wan, ali2025world, agarwal2026cosmos}, WAMs jointly reason about robot actions and their potential consequences, providing richer supervision for learning robot–environment interactions.

\begin{figure}[!t]
    \centering
    \includegraphics[width=0.95\columnwidth]{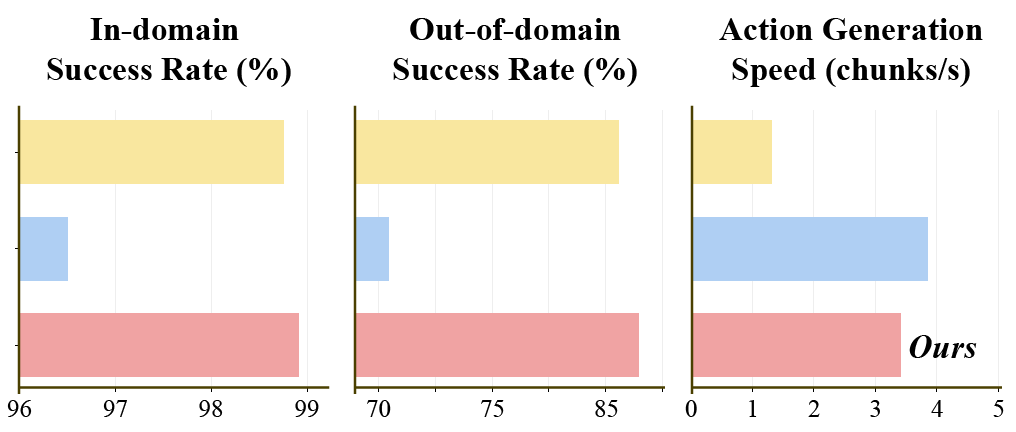}
    \caption{Comparison of Joint-WAM, Fast-WAM, and MoWAM, shown in yellow, blue, and red, respectively. In-domain and out-of-domain denote average success rates on LIBERO and LIBERO-Plus. Speed denotes the number of action chunks generated per second based on measured latency.}
    \label{fig:compare}
    \vspace{-12pt}
\end{figure}

Existing WAMs differ substantially in how future dynamics are used at inference time. One line of work ~\cite{zhu2025unified, Bi_2026_CVPR, ye2026world, li2026causal} explicitly generates visual scene futures, either by first generating future videos and conditioning action prediction on them, or by jointly generating future visual states and robot actions. Such explicit future prediction instantiates the learned dynamics under the current observation, providing an estimate of how the scene may evolve as the robot interacts with it. This can be particularly valuable under distribution shifts,  where explicitly estimating future scene evolution can help handle unseen situations. However, obtaining these benefits requires dense future video generation, which introduces substantial computational and memory overhead. At the same time, such predictions contain visual details that are not necessarily required for robot execution, such as background and texture information. A more efficient alternative ~\cite{yuan2026fast} removes explicit future generation at inference and predicts actions from observation features produced by a future video prediction model trained on robot task data. As a result, future dynamics information remains implicit in these observation features rather than being explicitly instantiated for the current scene. Our results suggest that this distinction becomes important under distribution shift. Implicit use of future dynamics remains effective in-distribution but degrades more noticeably on OOD observations, as shown in Fig.~\ref{fig:compare}. This may be due to the lack of information from an explicit future estimate tailored to the current scene. This observation motivates a central question: \textit{Can a WAM retain explicit future prediction at inference without paying the cost of generating complete future videos?}

To address this question, we propose MoWAM, which replaces dense future video generation at inference with explicit future motion prediction. Our key insight is that, for robot manipulation, the control-relevant information in a predicted future is largely reflected in how the robot is expected to move while interacting with the environment, rather than in the complete appearance of the future scene. MoWAM therefore represents the future using structured end-effector motion, capturing both spatial displacement and changes in gripper geometry as a compact abstraction of the future. To retain the benefits of learning future visual dynamics, MoWAM employs a Mixture-of-Transformer architecture consisting of a Video Transformer and an Action-Motion Transformer. The video Transformer learns future visual dynamics during training, while the Action-Motion Transformer jointly predicts robot actions and their corresponding future motion from the observation representation. Importantly, prediction of motion and action does not depend on generated future frames, allowing future video generation to be removed at inference while preserving an explicit prediction of future interaction dynamics. Because predicting motion is substantially cheaper than generating complete future videos, its compact representation makes inference-time scaling practical. Given the same observation and instruction, MoWAM can sample multiple candidates, each consisting of an action and its predicted future motion. We introduce a motion-aware task-progress verifier that evaluates these candidates based on their predicted future motion and executes the highest-scoring one. In this way, MoWAM can explore multiple explicit futures under a practical inference budget, while motion serves not only as a future representation for action generation but also as a structured signal for comparing the future consequence of different actions.

We evaluate MoWAM on the standard LIBERO ~\cite{liu2023libero} benchmark, the distribution-shifted LIBERO-Plus benchmark ~\cite{fei2025libero}, and real-world manipulation tasks. MoWAM achieves strong in-distribution performance while showing more noticeable gains under distribution shift, where explicit future prediction is particularly beneficial. In real-world experiments, it also improves average task success over representative WAM baselines while avoiding dense future video generation at inference. Moreover, increasing the candidates number consistently improves performance, demonstrating that compact future motion provides an effective mechanism for inference-time scaling.

Our contributions are threefold:
\begin{itemize}
        \item We introduce explicit future motion as an efficient future representation for WAMs, preserving explicit future reasoning without requiring dense future video generation at inference.
        \item We propose MoWAM, a world action model with explicit future motion prediction, together with a motion-based inference-time scaling strategy that generates and verifies multiple candidates of motion and action pairs to improve policy performance with additional test-time computation.
        \item We conduct extensive experiments on LIBERO, LIBERO-Plus, and real-world manipulation tasks, demonstrating strong in-distribution performance, improved OOD robustness, and better real-world success. We further show that increasing candidates of action and motion pairs consistently improves performance, validating MoWAM's inference-time scaling.
\end{itemize}

\section{RELATED WORK}

\subsection{World Action Models}

Vision-Language-Action (VLA) models ~\cite{brohan2023rt2, intelligence2025pi05, kim2024openvla, kim2025fine, pertsch2025fast, wang2026unified, team2024octo}  have emerged as a dominant paradigm for generalist robot control, leveraging large-scale vision-language priors to directly predict robot actions from observations and instructions. However, they do not explicitly model future environment dynamics. Recent advances in video foundation models \cite{wan2025wan, ali2025world} have motivated World Action Models (WAMs), which incorporate future dynamics into policy learning. Early efforts such as UWM~\cite{zhu2025unified} unify video and action diffusion within a shared architecture, supporting policy, forward-dynamics, and inverse-dynamics modeling. Motus~\cite{Bi_2026_CVPR} further unifies video generation, action prediction, and multimodal understanding through a latent action world model. Cosmos Policy~\cite{kim2026cosmos} encodes robot actions as latent frames and generates them together with future visual states within the latent space. DreamZero~\cite{ye2026world} builds upon a large pretrained video diffusion model to jointly model future video and actions, leveraging its strong priors for robot control. LingBot-VA~\cite{li2026causal} instead adopts an autoregressive formulation to predict future video and actions from their histories. Together, these works demonstrate the effectiveness of future-aware video-action modeling for robot control.

\subsection{Efficient Inference of World Action Models}

Future prediction can provide policy generation with additional information about scene evolution, but brings substantial inference overhead. To reduce inference cost of WAMs, several works improve WAM efficiency by directly reducing the computational cost of visual world modeling itself, through generative distillation~\cite{akbari2026flash}, coarse future modeling~\cite{li2026efficient}, or asynchronous world-model execution~\cite{cai2026aha}. More fundamentally, recent studies have reconsidered whether complete visual futures need to be generated at inference. Fast-WAM~\cite{yuan2026fast} retains future video prediction during training but removes explicit future generation at inference, using observation representations to predict actions. JEPA-WAM ~\cite{lin2026jepa} models future transitions in a pretrained joint-embedding space, coupling latent future prediction with action generation. LaWAM~\cite{chen2026lawam} retains future modeling at inference, but conditions action generation on future visual subgoals. These studies suggest that retaining the full future is not strictly required for action generation. MoWAM further explores structured robot motion as a compact and explicit representation of the future, providing an alternative to either removing future prediction at inference or representing the future only in latent spaces.

\section{METHOD}

\subsection{Problem Definition}
WAMs introduce future prediction to support action prediction. During training, future video prediction allows the model to learn how the scene evolves under robot actions. At inference, some WAMs generate a dense future visual rollout from the current observation to support action prediction. Given the current observation $\mathbf o$, language instruction $\ell$, and robot state $\mathbf s$, let $\mathbf z_o$ denote the observation latent, $\mathbf v$ the future visual latents, and $\mathbf a_{1:H}$ an action chunk with horizon $H$. The resulting action distribution can be written as:
\begin{equation}
p_{\theta,\phi}\left(
\mathbf a_{1:H}
\mid
\mathbf z_o,\ell,\mathbf s
\right)
=
\int
p_{\theta,\phi}\left(
\mathbf a_{1:H},\mathbf v
\mid
\mathbf z_o,\ell,\mathbf s
\right)
\,\mathrm d\mathbf v .
\label{eq:wam-action-marginalization}
\end{equation}
Here, $\theta$ and $\phi$ denote the parameters associated with visual future modeling and action modeling, respectively. Existing WAMs ~\cite{Bi_2026_CVPR, li2026causal, ye2026world} typically realize this formulation in two ways: future video and actions are either jointly modeled within a unified model with shared parameters, or future video is first generated by a video model and then used to condition action prediction. In both cases, inference involves a dense future visual rollout, introducing additional inference overhead. To reduce this, Fast-WAM~\cite{yuan2026fast} retains future video modeling in training but removes explicit future rollout generation at inference time. It directly predicts actions from the hidden representation of the observation $\mathbf h_{\mathrm{obv}}^\theta$ produced by the trained video model parameterized by $\theta$:
\begin{equation}
\mathbf a_{1:H}
\sim
p_\phi\left(
\mathbf a_{1:H}
\mid
\mathbf h_{\mathrm{obv}}^\theta,\ell,\mathbf s
\right).
\label{eq:fast-wam}
\end{equation}
The model maintains in-domain performance, but falls behind WAMs with explicit visual rollouts under out-of-distribution settings. This suggests explicit visual rollout at inference can still provide additional information for action prediction, particularly under distribution shift.

This raises a natural question: can we preserve this inference-time benefit without generating a dense future visual rollout? To this end, we use structured robot motion as a compact explicit representation of the future consequence of an action.  Rather than representing the control commands themselves, motion describes how the end effector is expected to evolve under the current scene and interaction constraints. We extend the action model $p_{\phi}$ to jointly model motion and action:
\begin{equation}
(\mathbf a_{1:H}, \mathbf m_{1:H})
\sim
p_\phi\left(
\mathbf a_{1:H}, \mathbf m_{1:H}
\mid
\mathbf h_{\mathrm{obv}}^\theta, \ell, \mathbf s
\right),
\label{eq:MoWAM}
\end{equation}
where $\mathbf m_{1:H}$ denotes the predicted structured robot motion. During training, the model retains future video modeling and jointly learns motion and action. At inference time, it generates only motion and action without generating future video.

\subsection{Structured Motion Representation}

\begin{figure*}[!t]
  \centering
  \includegraphics[width=0.87\textwidth]{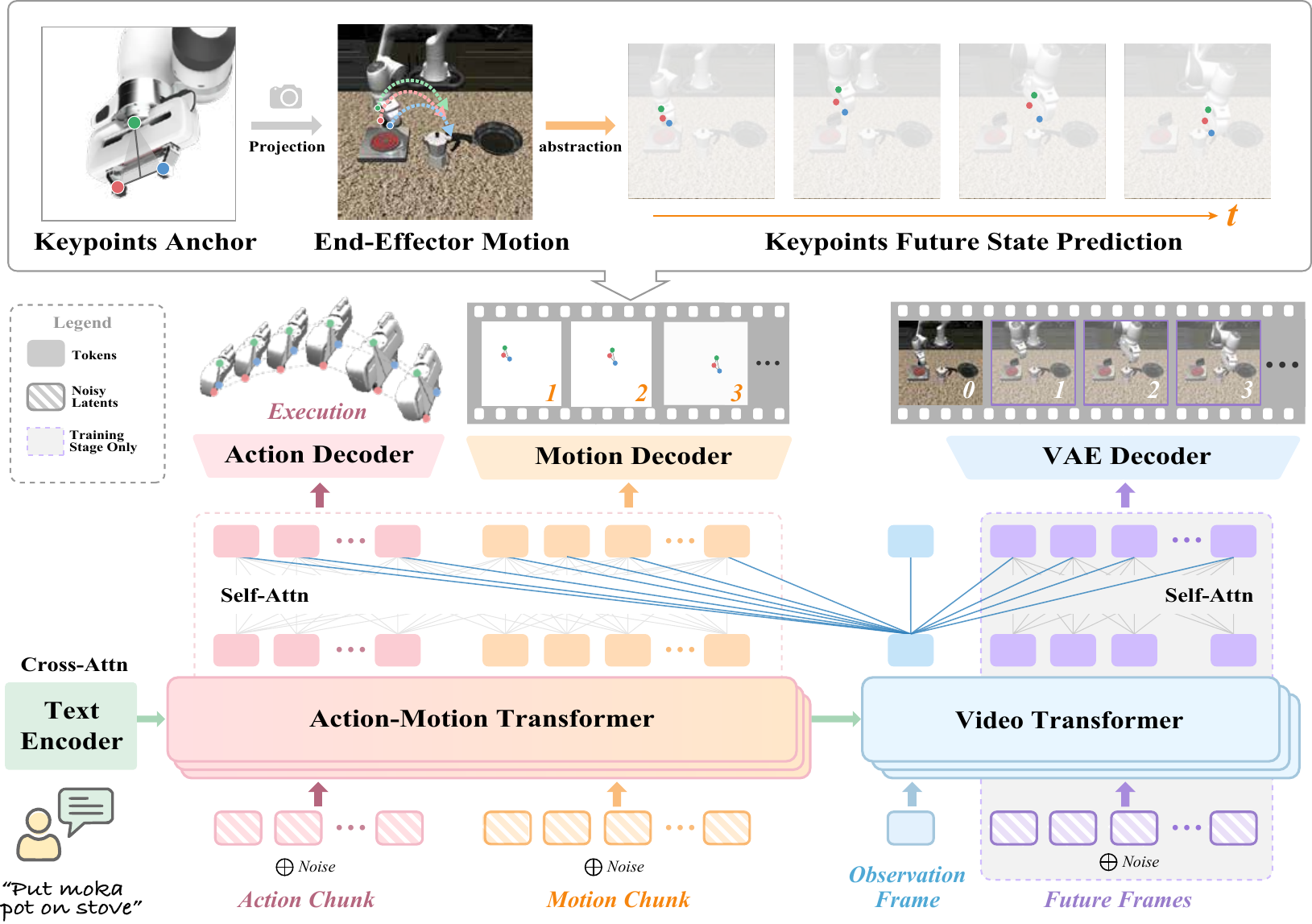}
  \caption{Framework of MoWAM. The Video Transformer learns future dynamics during training, while the Action-Motion Transformer predicts action and motion without future video generation at inference. Structured end-effector motion provides a compact abstraction of the future, replacing dense future visual rollout at inference.} 
  \label{fig:motionscaling-framework}
\vspace{-10pt}
\end{figure*}

At inference time, explicitly generating future videos in WAMs introduces substantial computational overhead, while directly predicting actions from the observation features loses the additional information provided by explicit future prediction. Under a given action, the future interaction is reflected in how the end effector moves and changes its state under the current scene and task constraints. We characterize this evolution using a small set of fingertip and palm keypoints. Their joint displacement captures overall end-effector movement, while changes in their relative geometry reflect variations in gripper state, including opening and closing. We therefore use these quantities to form structured end-effector motion, retaining the robot's future movement and changes in gripper geometry while discarding dense visual details of the future scene. Specifically, for a future time step \(t\) in camera view \(j\), we denote
$\mathbf p_{t,l}^j,\mathbf p_{t,r}^j,\mathbf p_{t,p}^j\in\mathbb R^2$
as the projected positions in the image of the left fingertip, right fingertip, and palm keypoints, respectively, which are selected according to the end-effector configuration. Based on these, we define the gripper center \(\mathbf c_t^j\), fingertip span \(\mathbf s_t^j\), and relative palm offset \(\mathbf r_t^j\) as:
\[
\left\{
\begin{aligned}
\mathbf s_t^j &=
\mathbf p_{t,r}^j-\mathbf p_{t,l}^j,\\
\mathbf c_t^j &=
\frac{\mathbf p_{t,l}^j+\mathbf p_{t,r}^j}{2},\\
\mathbf r_t^j &=
\mathbf p_{t,p}^j-\mathbf c_t^j.
\end{aligned}
\right.
\]
As illustrated in Fig.~\ref{fig:motionscaling-framework}, we characterize end-effector motion by the changes in these three geometric quantities between consecutive time steps.  Accordingly, we define motion at time step $t$ as: 
\begin{equation}
\mathbf m_{\mathrm{dyn},t}^j =
\begin{bmatrix}
\mathbf c_t^j-\mathbf c_{t-1}^j\\
\mathbf s_t^j-\mathbf s_{t-1}^j\\
\mathbf r_t^j-\mathbf r_{t-1}^j
\end{bmatrix}
\in \mathbb{R}^6,\quad
t=1,\ldots,H.
\end{equation}
Importantly, motion is not obtained by directly projecting action commands into image. Instead, it predicts how the robot is expected to move when the action is carried out in the current scene.  In addition to temporal motion, motion should preserve the geometric structure of the end effector. We therefore define the relative-geometry component using the fingertip span vector $\mathbf s_t^j$ and relative palm offset $\mathbf r_t^j$ at each future time step:
\begin{equation}
\mathbf m_{\mathrm{str},t}^{j}
=
\begin{bmatrix}
\mathbf s_t^j\\
\mathbf r_t^j
\end{bmatrix}
\in\mathbb R^4,\qquad t=1,\ldots,H.
\end{equation}
Together, these two components define the structured robot motion, denoted as $\mathbf m=(\mathbf m_{\mathrm{dyn}},\mathbf m_{\mathrm{str}})$, where $\mathbf m_{\mathrm{dyn}}$ and $\mathbf m_{\mathrm{str}}$ collect the corresponding quantities across time steps and camera views.

\subsection{MoWAM Framework}
MoWAM retains future video modeling during training to learn scene dynamics, while decoupling generation of action and motion from future video prediction. This allows future video generation to be removed at inference. Specifically, MoWAM adopts a dual-branch Transformer architecture consisting of a Video Transformer and an Action-Motion Transformer. The Video Transformer learns future visual dynamics through future video prediction, whereas the Action-Motion Transformer jointly models robot action and future motion. Both branches are conditioned on the current observation and language instruction, while generation of action and motion does not depend on generated future frames. In this way, future video prediction supports dynamics learning during training, while the Action-Motion branch can independently generate robot actions at inference. Motion is represented by the geometric quantities introduced in Sec. B, capturing both the robot's overall movement and local geometric changes of the end effector.

In training, the observed frames are first encoded into latent tokens $\mathbf z_{\mathrm{obv}} = E_{\mathrm{VAE}}(\mathbf O)$. The future video frames, action chunk $\mathbf a=\mathbf a_{1:H}$, and structured motion $\mathbf m$ are then processed by their corresponding encoders and added with noise. As illustrated in Fig. ~\ref{fig:motionscaling-framework}, for future video modeling, the Video Transformer $G_\theta$ learns to denoise the noisy future video frames conditioned on the current observation and language instruction, without receiving information from the Action–Motion branch. Actions and structured motion are jointly modeled in the Action–Motion Transformer $G_\phi$, where action and motion tokens access the observation representation $\mathbf h_{\mathrm{obv}}^\theta$ through the corresponding keys and values in attention. We denote these observation keys and values as
$$
\mathrm{KV}_{\mathrm{obv}}^{\theta}
=
\left(
\mathbf K_{\mathrm{obv}}^{\theta},
\mathbf V_{\mathrm{obv}}^{\theta}
\right),
$$

Then, the attention in the Action-Motion Transformer can be formulated as:

$$
\mathbf H_{am}^{\prime}
=
\operatorname{Attn}
\left(
\mathbf Q_{am},
\left[
\mathbf K_{\mathrm{obv}}^{\theta};
\mathbf K_{am}
\right],
\left[
\mathbf V_{\mathrm{obv}}^{\theta};
\mathbf V_{am}
\right]
\right),
$$
Here, $\mathbf Q_{am}$, $\mathbf K_{am}$, and $\mathbf V_{am}$ are obtained by projecting the action and motion tokens. In this way, action and motion exchange information within the same attention computation while directly accessing the current observation representation. This allows generation of action and motion to leverage the dynamics learned by the Video Transformer $G_\theta$ through future video prediction. The language instruction is incorporated separately through cross-attention in the two Transformer branches.
Since the same observation and task instruction allow multiple plausible future robot motions, we model motion as a conditional distribution using flow matching~\cite{lipman2022flow}. Motion supervision consists of two components, corresponding to the temporal changes $\mathbf m_{\mathrm{dyn}}$ and relative geometry $\mathbf m_{\mathrm{str}}$ defined in Sec. B. Accordingly, the conditional distributions parameterized by the Video and Action-Motion Transformers are denoted as $p_\theta$ and $p_\phi$, and the training-time formulation of MoWAM can be written as
$$
p_\theta
\left(
\mathbf v
\mid
\mathbf z_{\mathrm{obv}},\ell
\right)
p_\phi
\left(
\mathbf a_{1:H},\mathbf m_{1:H}
\mid
\mathbf h_{\mathrm{obv}}^\theta,\ell,\mathbf s
\right),
$$

At inference time, the current observation is passed through the Video Transformer $G_\theta$ only once to obtain $\mathrm{KV}{\mathrm{obv}}^\theta$, without further generating future videos. Action and motion are initialized from random noise and jointly denoised in the Action–Motion Transformer $_\phi$, conditioned on $\mathrm{KV}{\mathrm{obv}}^\theta$, to recover the action chunk $\mathbf a_{1:H}$ and its corresponding future motion. Finally, the action chunk is decoded by the Action Decoder into executable robot control commands.

\subsection{Motion-Based Inference-Time Scaling}
To select among candidate motions under a given observation and task instruction, we learn a conditional scoring function $s_\psi(I,\ell,\mathbf m)$, where $I$, $\ell$, and $\mathbf m$ denote the current image, language instruction, and candidate future robot motion, respectively. We construct ranking supervision from successful demonstrations by treating the demonstrated motion $\mathbf m^+$ as preferable to its perturbed variants $\mathbf m^-$. Negative samples are generated while keeping the initial robot state unchanged by reducing motion changes, adding a smooth spatial shift, or cutting off the later part of the motion to simulate early stopping. These perturbations introduce differences in motion shape, final position, and completeness. We train the verifier using a pairwise ranking loss:
$$
\mathcal{L}_{\mathrm{rank}} = \mathbb{E}\!\left[ \max\!\left( 0,\, \gamma -s_\psi(I,\ell,\mathbf m^+) +s_\psi(I,\ell,\mathbf m^-) \right) \right]
$$
where $\gamma$ is a fixed ranking margin. 

At inference time, the verifier scores all generated candidates and the action associated with the highest-scoring candidate is executed. 

\begin{figure}[!b]
    \centering
    \vspace{-6pt}
    \includegraphics[width=0.99\columnwidth]{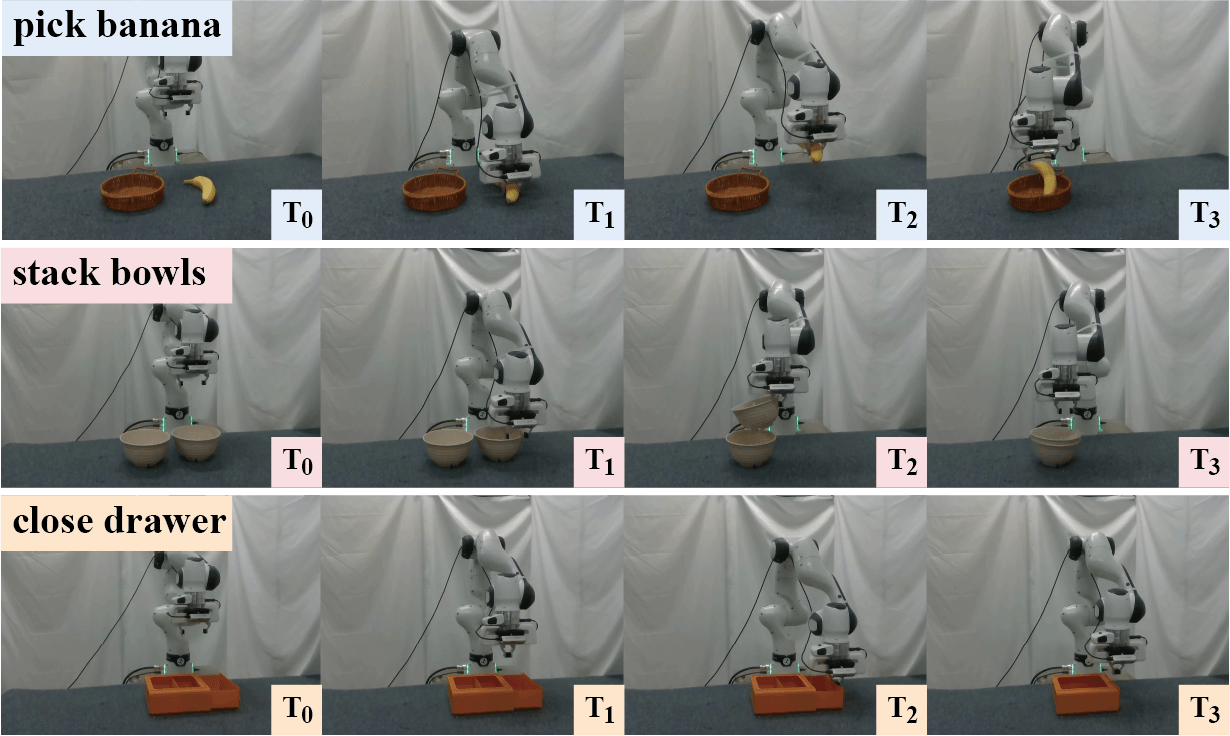}
    \caption{Three real-world tasks on a Franka Research 3 robot: Pick Banana, Stack Bowls, and Close Drawer}
    \label{fig:demo}
\end{figure}

\begin{table*}[!t]
\begin{minipage}[t]{\columnwidth}
\centering
\caption{Comparison of task success rates (\%) on the LIBERO benchmark across four task suites. Robo P.T. indicates whether robotic embodied pretraining is used. The highest and second-highest results in each column are highlighted in bold and underlined, respectively.}
\label{tab:libero}
\renewcommand{\arraystretch}{1.0}
\setlength{\tabcolsep}{3.9pt}

\begin{tabular}{lcSMMMA}
\toprule
\textbf{Method} & \textbf{Robo. P.T.} & \textbf{Spa.} & \textbf{Obj.} & \textbf{Goa.} & \textbf{Lon.} & \textbf{Avg.} \\[1.5pt]
\specialrule{\lightrulewidth}{0pt}{0pt}
\rowcolor{VLABlue}
\multicolumn{7}{c}{%
    \rule[-0.55ex]{0pt}{2.9ex}\textit{VLA-based Methods}
} \\
\specialrule{\lightrulewidth}{0pt}{0pt}
\addlinespace[2.5pt]

OpenVLA~\cite{kim2024openvla}      & $\checkmark$ & 84.7 & 88.4 & 79.2 & 53.7 & 76.5 \\[4pt]
OpenVLA-OFT~\cite{kim2025fine}  & $\checkmark$ & 97.2 & 97.8 & 96 & 96 & 96.75 \\[4pt]
UniVLA~\cite{wang2026unified}       & $\checkmark$ & 95.4 & 94.8 & 94.8 & 90.8 & 93.95 \\[4pt]
$\pi_0$~\cite{black2024pi_0}      & $\checkmark$ & 96.8 & 98.8 & 95.8 & 85.2 & 94.1 \\[4pt]
$\pi_0$-FAST~\cite{pertsch2025fast} & $\checkmark$ & 97.8 & 97.8 & 88.2 & 61 & 86.2 \\[4pt]
$\pi_{0.5}$~\cite{intelligence2025pi05}  & $\checkmark$ & \underline{98.8} & 98.2 & 98 & 92.4 & 96.9 \\[3pt]

\specialrule{\lightrulewidth}{0pt}{0pt}
\rowcolor{WAMGreen}
\multicolumn{7}{c}{%
    \rule[-0.55ex]{0pt}{2.9ex}\textit{ WAM-based Methods}
} \\
\specialrule{\lightrulewidth}{0pt}{0pt}
\addlinespace[2.5pt]

Lingbot-VA~\cite{li2026causal} & $\checkmark$ & 98.5 & \underline{99.6} & 97.2 & \textbf{98.5} & 98.5 \\[4pt]
Motus~\cite{Bi_2026_CVPR}      & $\checkmark$ & 96.8 & \textbf{99.8} & 96.6 & 97.6 & 97.7 \\[4pt]
LaWAM~\cite{chen2026lawam}      & $\checkmark$ & \textbf{99} & 96 & 97.2 & 93.4 & 96.4 \\[4pt]
Fast-WAM~\cite{yuan2026fast}   & $\times$     & 96.6 & 99.2 & 94.6 & 95.6 & 96.5 \\[4pt]
IDM-WAM    & $\times$     & 98.6 & 99.4 & \underline{98.4} & 96.4 & 98.2 \\[4pt]
Joint-WAM  & $\times$     & \textbf{99} & 99.2 & \textbf{99} & \underline{97.8} & \underline{98.75} \\[3pt]

\specialrule{\lightrulewidth}{0pt}{0pt}
\rowcolor{gray!8}
\rule[-0.95ex]{0pt}{3.4ex}\textbf{Ours}
& $\times$
& \underline{98.8}
& \textbf{99.8}
& \textbf{99}
& \underline{97.8}
& \textbf{98.9} \\
\specialrule{\heavyrulewidth}{0pt}{0pt}

\end{tabular}
\end{minipage}\hfill
\begin{minipage}[t]{\columnwidth}

\centering
\caption{Comparison with two representative methods on three real-world tasks in success rate (\%) and action latency ($\downarrow$). Robo P.T. denotes embodied pretraining. Best and second-best averages are bolded and underlined.}
\label{tab:real_exper}
\label{tab:real_exper}
\renewcommand{\arraystretch}{1.0}
\setlength{\tabcolsep}{3.2pt}

\begin{tabularx}{0.98\columnwidth}{l c Y Y Y Y Y}
\toprule

\begin{tabular}[c]{@{}l@{}}\textbf{Method}\end{tabular}
&
\begin{tabular}[c]{@{}c@{}}\textbf{Robo. P.T.}\end{tabular}
&
\begin{tabular}[c]{@{}c@{}}\textbf{Pick}\\\textbf{Banana}\end{tabular}
&
\begin{tabular}[c]{@{}c@{}}\textbf{Stack}\\\textbf{Bowls}\end{tabular}
&
\begin{tabular}[c]{@{}c@{}}\textbf{Close}\\\textbf{Drawer}\end{tabular}
&
\begin{tabular}[c]{@{}c@{}}\textbf{Avg.}\end{tabular}
&
\begin{tabular}[c]{@{}c@{}}\textbf{Action}\\\textbf{Latency}\end{tabular}
\\[1.5pt]

\midrule
\addlinespace[2.5pt]

\rule[-0.45ex]{0pt}{2.7ex}Motus~\cite{Bi_2026_CVPR}
& $\checkmark$
& \textbf{65}
& 65
& \textbf{85}
& \underline{71.67}
& 1621.9 \\[2pt]

\rule[-0.45ex]{0pt}{2.7ex}Fast-WAM~\cite{yuan2026fast}
& $\times$
& \underline{30}
& \underline{85}
& 20
& 45
& \textbf{260.1} \\[2pt]

\specialrule{\lightrulewidth}{0pt}{0pt}
\rowcolor{gray!8}
\rule[-0.95ex]{0pt}{3.4ex}\textbf{Ours}
& $\times$
& \textbf{65}
& \textbf{95}
& \underline{80}
& \textbf{80}
& \underline{293.5} \\

\specialrule{\heavyrulewidth}{0pt}{0pt}

\end{tabularx}
\par\vspace{\floatsep}

\begingroup
    \centering
    \includegraphics[width=\columnwidth]{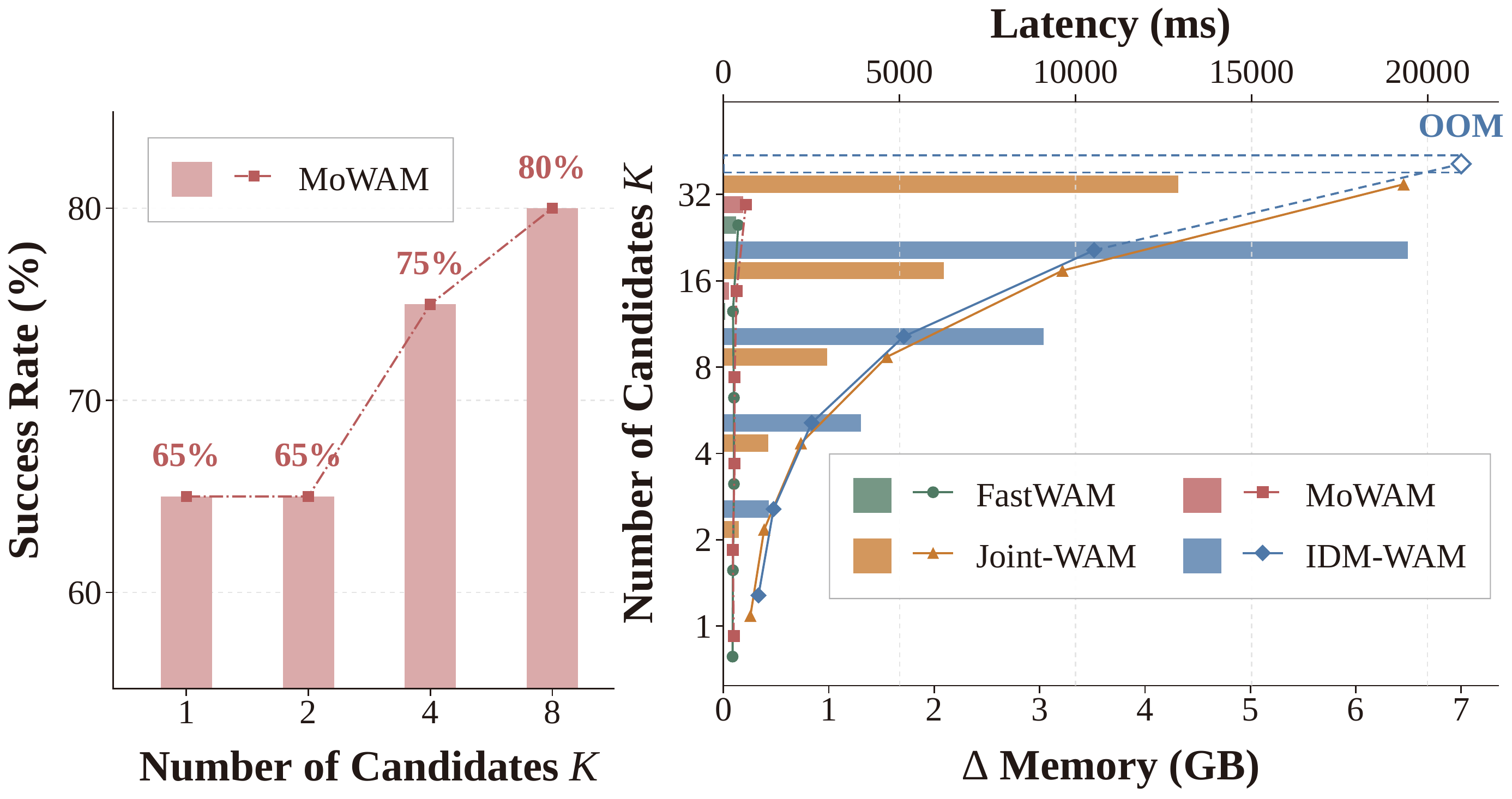}
    \captionof{figure}{Left: increasing the number of candidates improves task success rate. Right: latency and memory consumption scale with the number of candidates for different WAM variants.}
    \label{fig:scaling}
\endgroup
\end{minipage}
\par\vspace{\floatsep}

\centering
\caption{Comparison on LIBERO-Plus across seven OOD settings, measured by task success rate (\%). Embodied P.T. denotes embodied pretraining. The best and second-best results in each column are marked in bold and underlined, respectively.}
\label{tab:liberoplus}
\renewcommand{\arraystretch}{1.0}
\setlength{\tabcolsep}{3.6pt}

\begin{tabularx}{0.995\textwidth}{l c Y Y Y Y Y Y Y Y Y}
\toprule
\textbf{Method} & \textbf{Robo. P.T.} & \textbf{Objects} & \textbf{Camera} & \textbf{Initial} & \textbf{Light} & \textbf{Background} & \textbf{Sensor} & \textbf{Language} & \textbf{Average} & \textbf{Latency} \\[1.5pt]
\midrule
\addlinespace[2.5pt]

OpenVLA~\cite{kim2024openvla}      & $\checkmark$ & 53 & 2 & 15 & 24 & 54 & 32 & 44 & 32.00 & 128.71 \\[4pt]
OpenVLA-OFT~\cite{kim2025fine}  & $\checkmark$ & 73 & \textbf{65} & 27 & 94 & 89 & 70 & 87 & 72.14 & \underline{103.3} \\[4pt]
UniVLA~\cite{wang2026unified}       & $\checkmark$ & 76 & 8 & 62 & 77 & \underline{90} & 19 & 80 & 58.86 & 492.2 \\[4pt]
$\pi_0$~\cite{black2024pi_0}      & $\checkmark$ & 81 & \underline{62} & 40 & 92 & 81 & 75 & 70 & 71.57 & \textbf{62.9} \\[4pt]
$\pi_0$-FAST~\cite{pertsch2025fast} & $\checkmark$ & 73 & \underline{62} & 26 & 72 & 76 & 70 & 78 & 65.29 & 274.22 \\[4pt]
Motus~\cite{Bi_2026_CVPR}        & $\checkmark$ & \textbf{89} & 48 & \underline{88} & 83 & 78 & 57 & 89 & 76.00 & 1621.9 \\[4pt]
LaWAM~\cite{chen2026lawam}        & $\checkmark$ & 84 & 37 & 70 & 92 & \textbf{95} & 73 & \textbf{98} & 78.43 & 108.52 \\[4pt]
Fast-WAM~\cite{yuan2026fast}     & $\times$     & 82 & 42 & 74 & 85 & 63 & 72 & 75 & 70.43 & 260.1 \\[4pt]
IDM-WAM      & $\times$     & 84 & 59 & 79 & 90 & 68 & \textbf{86} & 95 & 80.14 & 995.4 \\[4pt]
Joint-WAM    & $\times$     & 85 & 51 & \textbf{91} & \underline{95} & 67 & \underline{78} & \underline{97} & \underline{80.57} & 766.3 \\[3pt]

\specialrule{\lightrulewidth}{0pt}{0pt}
\rowcolor{gray!8}
\rule[-0.95ex]{0pt}{3.4ex}\textbf{Ours} & $\times$ & \underline{87} & 51 & 86 & \textbf{96} & 68 & \textbf{86} & 96 & \textbf{81.43} & 293.5 \\
\specialrule{\heavyrulewidth}{0pt}{0pt}
\end{tabularx}
\vspace{-6pt}
\end{table*}

\section{EXPERIMENT}

\subsection{Experimental Setup}
\subsubsection{Benchmark}
We evaluate MoWAM and other baselines on the standard LIBERO benchmark~\cite{liu2023libero}, the out-of-distribution benchmark LIBERO-Plus~\cite{fei2025libero}, and real-world single-arm manipulation tasks, using task success rate as the primary evaluation metric.

\textbf{LIBERO.} LIBERO is a standard benchmark for language-conditioned robotic manipulation, comprising four task suites. Each suite contains 10 tasks and provides 500 demonstration trajectories. We perform 50 rollouts for each task and report the task success rate.

\textbf{LIBERO-Plus.} LIBERO-Plus provides a systematic evaluation of policy robustness under out-of-distribution conditions. The benchmark includes seven types of distribution shifts. For each perturbation type, we uniformly sample 100 evaluation instances from the four LIBERO task suites, resulting in a total of 700 OOD evaluation tasks. All methods are evaluated directly using policies trained on LIBERO, without any additional training or adaptation. For the ablation study, all variants are trained exclusively on the LIBERO-Long data and evaluated on the corresponding LIBERO-Plus settings derived from LIBERO-Long.

\textbf{Real-World Evaluation.} We conduct real-world experiments on a Franka Research 3 robot. We consider three representative manipulation tasks: Pick Banana, Close Drawer, and Stack Bowls, as shown in Fig.~\ref{fig:demo}. These tasks require stable grasping, accurate spatial positioning, and controlled interaction with the environment, covering different real-world manipulation behaviors and interaction requirements. We collect 100 demonstrations for each task and conduct 20 evaluation trials per task under the standard setting.

\subsubsection{Implementation Details}
We use pretrained Wan2.2-5B~\cite{wan2025wan}. The Action-Motion Transformer follows the same architecture as the Video Transformer with a hidden dimension of 1024. We largely follow experimental settings of Fast-WAM~\cite{yuan2026fast}, setting both the action and motion horizons to 32 and temporally downsampling videos by $4\times$ to nine frames per chunk. All models are trained for 20000 steps on four NVIDIA A100 GPUs with jointly sampled multi-task data. Fast-WAM, Joint-WAM, and IDM-WAM use the same training configuration and model architecture, differing only in their future-modeling strategies. At inference, we use 10 denoising steps with CFG=1.0. All speed and latency measurements are conducted on a single NVIDIA RTX 5090 GPU, with identical action horizons.

\begin{table*}[t]
\centering
\label{tab:ablation_liberoplus}
\caption{Ablation studies of MoWAM on LIBERO-Long of LIBERO-Plus. (1) Motion supervision ablates the dynamics and structure supervision for motion learning. (2) Training time video modeling freezes the Video DiT to assess the contribution of future video modeling. (3) Motion conditioning ablation prevents action from attending to motion.}
\renewcommand{\arraystretch}{1.0}
\setlength{\tabcolsep}{4pt}

\begin{tabularx}{0.975\textwidth}{>{\raggedright\arraybackslash}p{3.2cm} Y Y Y Y Y Y Y >{\raggedright\arraybackslash}p{1.555cm}}
\toprule
\textbf{Method} & \textbf{Objects} & \textbf{Camera} & \textbf{Initial} & \textbf{Light}
& \textbf{Background} & \textbf{Sensor} & \textbf{Language} &  \multicolumn{1}{c}{\textbf{Average SR}} \\[1.5pt]

\specialrule{\lightrulewidth}{0pt}{0pt}
\rowcolor{VLABlue}
\multicolumn{9}{c}{\rule[-0.99ex]{0pt}{3.5ex} \textbf{\textit{1) Motion Supervision Ablation}}} \\
\specialrule{\lightrulewidth}{0pt}{0pt}
\addlinespace[2.5pt]

w/o Motion           & 92 & 32 & 84 & 88 & 32 & 76 & 56 & 65.71 \\[3pt]
w/o Dynamics Loss      & 80 & 36 & 92 & 84 & 52 & 68 & 72 & 69.14 {\color{DeepRed}(3.43$\uparrow$)} \\[3pt]
w/o Structure Loss   & 88 & 48 & 84 & 76 & 52 & 76 & 92 & 73.71 {\color{DeepRed}(8$\uparrow$)} \\[3pt]
Ours                 & 88 & 36 & 92 & 84 & 52 & 88 & 92 & \textbf{76 {\color{DeepRed}(10.29$\uparrow$)}} \\[3pt]

\specialrule{\lightrulewidth}{0pt}{0pt}
\rowcolor{WAMGreen}
\multicolumn{9}{c}{\rule[-0.99ex]{0pt}{3.5ex} \textbf{\textit{2) Training-Time Video Modeling Ablation}}} \\
\specialrule{\lightrulewidth}{0pt}{0pt}
\addlinespace[2.5pt]
w/o Video DiT Training & 20 & 32 & 56 & 52 & 4 & 40 & 28 & 33.14 \\[3pt]
w/o Motion           & 92 & 32 & 84 & 88 & 32 & 76 & 56 & 65.71 \\[3pt]
Ours                   & 88 & 36 & 92 & 84 & 52 & 88 & 92 & \textbf{76 {\color{DeepRed}(42.86$\uparrow$)}} \\[3pt]

\specialrule{\lightrulewidth}{0pt}{0pt}
\rowcolor{AblYellow}
\multicolumn{9}{c}{\rule[-0.99ex]{0pt}{3.5ex} \textbf{\textit{3) Motion Conditioning Ablation}}} \\
\specialrule{\lightrulewidth}{0pt}{0pt}
\addlinespace[2.5pt]

w/o Motion           & 92 & 32 & 84 & 88 & 32 & 76 & 56 & 65.71 \\[3pt]
w/o Motion Conditioning & 84 & 40 & 92 & 84 & 48 & 72 & 76 & 70.86 {\color{DeepRed}(5.15$\uparrow$)} \\[3pt]
Ours                 & 88 & 36 & 92 & 84 & 52 & 88 & 92 & \textbf{76 {\color{DeepRed}(10.29$\uparrow$)}} \\

\bottomrule
\end{tabularx}
\vspace{-8pt}
\end{table*}


\subsection{Main Results}
\subsubsection{LIBERO}
Table~\ref{tab:libero} compares VLA- and WAM-based policies across the four LIBERO suites. MoWAM achieves an average success rate of 98.9\% without additional embodied pretraining. Compared with Fast-WAM, MoWAM improves the average success rate by 2.4 percentage points, showing that explicitly predicting future robot motion can further improve action generation. MoWAM also performs comparably to LingBot-VA, Motus, Joint-WAM, and IDM-WAM without generating dense future visual rollouts at inference, indicating that comparable in-domain performance can be retained using structured motion as the explicit future representation. Within compact future modeling, MoWAM obtains a 2.5-point higher average success rate than LaWAM. Together, these results indicate that our method performs well under in-domain settings.

\subsubsection{LIBERO-Plus}
As shown in Table~\ref{tab:liberoplus}, MoWAM achieves an average success rate of 81.43\% on LIBERO-Plus, outperforming the evaluated VLA baselines and exceeding the embodied-pretrained WAM Motus by 5.43 percentage points. MoWAM also slightly outperforms LaWAM, a compact latent-future method, by 3.0 percentage points on average. Under aligned training settings, MoWAM improves over Fast-WAM by 11.0 percentage points while adding only 33.4~ms of action generation latency. It achieves comparable average performance to baselines with dense future visual rollout in inference, including Joint-WAM and IDM-WAM, while reducing latency from 766.3~ms and 995.4~ms to 293.5~ms, respectively. The perturbation results of each type further show how much of the benefit of full visual-future generation can be retained by compact motion. On Initial and Sensor, MoWAM improves over Fast-WAM by 12 and 14 percentage points, respectively. On Light, Objects, and Background, it matches or exceeds both full visual-future baselines. These results indicate that compact motion can preserve the benefits of explicit future modeling across diverse OOD perturbations, rather than only under a particular type of distribution shift. However, this replacement is not uniformly equivalent to full visual-future generation: although MoWAM improves over Fast-WAM on both Initial and Camera, it remains 5 and 8 percentage points below the best visual-future results from Joint-WAM and IDM-WAM, respectively. Overall, these results support structured robot motion as a low-cost alternative to full visual-future generation, retaining comparable overall OOD performance.

\subsubsection{Real-World Evaluation}
Table~\ref{tab:real_exper} compares MoWAM with two representative baselines on three real-world manipulation tasks. MoWAM outperforms Fast-WAM on all three tasks, matches Motus on Pick Banana, leads by 30 points on Stack Bowls, and performs slightly worse on Close Drawer. Motus performs well on Pick Banana and Close Drawer but is less effective on Stack Bowls, whereas Fast-WAM shows the opposite trend. MoWAM achieves an average success rate of 80\%, indicating that the effectiveness of jointly modeling of motion and action extends beyond simulation to real-world tasks with different object relations and manipulation requirements. This performance is achieved with only a small additional inference cost: compared with Fast-WAM, MoWAM improves the average success rate by 35 percentage points while adding only 33.4~ms of action-generation latency. Compared with Motus, MoWAM reduces the latency from 1621.9~ms to 293.5~ms while achieving a higher average success rate. Overall, these results support structured robot motion as a compact future representation across different manipulation tasks, while keeping the inference cost close to that of a policy without explicit future generation.

\subsubsection{Inference Time Scaling}
Figure~\ref{fig:scaling} evaluates the performance gains and generation costs of increasing the number of candidates. As shown in part (a), increasing the candidate number from 1 to 4 and 8 raises the success rate on Pick Banana from 65\% to 75\% and 80\%, respectively. Since the verifier evaluates candidates based on their predicted Motion, these results show that the predicted motion can be used as a signal for candidate evaluation. To further examine the cost of such scaling under different future representations, we compare Fast-WAM, Joint-WAM, IDM-WAM, and MoWAM, which share the same architecture but differ in their future-modeling strategies. As shown in Fig.~\ref{fig:scaling}, MoWAM exhibits latency and memory growth close to Fast-WAM as the candidate count increases, whereas the cost of Joint-WAM and IDM-WAM grows substantially faster due to repeated full visual-future generation. This difference becomes increasingly pronounced at larger scales, with IDM-WAM running out of memory at $K=32$. Notably, MoWAM can generate eight candidates in less time than either full-visual-future variant requires for a single candidate. This suggests that computation saved by avoiding full visual future generation can instead be allocated to candidate exploration, which yields improved task success in our scaling experiment. These results show that compact motion makes inference-time candidate exploration substantially more practical than full visual-future generation.

\subsection{Ablation Study}
\subsubsection{Motion Supervision Ablation} We first examine the two supervision signals used to learn the structured motion representation. Specifically, w/o Dynamics Loss removes the supervision on motion changes while retaining Structure supervision, resulting in a decrease in the average success rate from 76.0\% to 69.14\%. This drop indicates that explicitly supervising temporal motion changes provides useful information for action prediction. Meanwhile, w/o Structure removes Structure supervision while retaining Dynamics supervision, reducing the average success rate to 73.71\%. This result suggests that Structure supervision provides complementary geometric information that benefits action prediction. Overall, these results support motion as an effective representation of future robot execution for action prediction.

\subsubsection{Training-Time Video Modeling Ablation}
We examine whether training-time future video modeling remains important when motion is used as the explicit future representation at inference. We freeze the Video DiT and use observation features produced by pretrained video model for policy generation. Despite retaining motion modeling, average success drops from 76.0\% to 33.1\%, compared with 65.7\% when future video training is retained without motion. This shows that learning scene evolution during training remains critical: future video prediction provides scene-level supervision for future visual dynamics and leverages pretrained video priors, whereas motion alone captures only future robot movement. Motion therefore serves a different role from future video training, providing a compact explicit future representation for action generation at inference.

\begin{table}[t]
\centering
\caption{Comparison between predicted motion and action-projected motion in trajectory prediction error.}
\label{tab:motion_ade}
\renewcommand{\arraystretch}{1.1}
\setlength{\tabcolsep}{7pt}

\begin{tabular}{lccc}
\toprule
\textbf{Setting} &
\textbf{Motion ADE} &
\textbf{Action-Proj. ADE} &
\textbf{ADE Reduction} \\
\midrule
ID  & 5.63 & 77.56 & 92.75\% \\
OOD & 9.38 & 63.86 & 85.31\% \\
\bottomrule
\end{tabular}
\vspace{-14pt}
\end{table}

\subsubsection{Motion Conditioning Ablation}
To determine whether the gains come from motion supervision itself or from conditioning action on predicted motion, we conduct a motion conditioning ablation. We retain motion prediction and supervision but prevent action tokens from attending to motion tokens. This variant achieves 70.86\% average success, compared with 65.71\% for w/o Motion setting, showing that motion supervision alone improves policy learning by providing signals on execution dynamics and relative geometry. Allowing action to additionally attend to motion further increases performance to 76\%, indicating that predicted motion also provides useful future information for action generation at inference.

\subsubsection{Accuracy of Future Motion Prediction}
We evaluate whether predicted motion captures the future execution of its paired action by comparing it with the actual execution trajectory in pixels on the LIBERO suite. For Action-Proj., we roll out the action sequence from the current state by converting each action into translation and rotation increments according to the simulator scale and project the resulting keypoints onto the image. As shown in Table~\ref{tab:motion_ade}, predicted motion achieves an ADE of 5.63 in ID and 9.38 in OOD, compared with 77.56 and 63.86 for Action-Proj., reducing the error by 92.75\% and 85.31\%, respectively. This shows that projected control commands do not accurately reflect the resulting robot motion, whereas predicted motion provides a substantially closer estimate of future execution. Although the error increases under distribution shift, this advantage remains.

\section{Conclusion}
We presented MoWAM, an efficient World Action Model that uses structured robot motion as a compact abstraction of the future. MoWAM retains future video modeling during training while avoiding dense future video generation at inference. Experiments on LIBERO, LIBERO-Plus, and real-world tasks demonstrate strong performance and improved robustness with low inference cost. Moreover, motion enables efficient inference-time scaling through multiple candidates of motion and action pairs.
\bibliographystyle{IEEEtran}
\bibliography{references}

@article{intelligence2025pi05,
  title   = {$\pi_{0.5}$: A Vision-Language-Action Model with Open-World Generalization},
  author  = {{Physical Intelligence}},
  journal = {arXiv preprint arXiv:2504.16054},
  year    = {2025}
}

@article{kim2024openvla,
  title   = {{OpenVLA}: An Open-Source Vision-Language-Action Model},
  author  = {Kim, Moo Jin and Pertsch, Karl and Karamcheti, Siddharth and Xiao, Ted and Balakrishna, Ashwin and Nair, Suraj and Rafailov, Rafael and Foster, Ethan and Lam, Grace and Sanketi, Pannag and Vuong, Quan and Kollar, Thomas and Burchfiel, Benjamin and Tedrake, Russ and Sadigh, Dorsa and Levine, Sergey and Liang, Percy and Finn, Chelsea},
  journal = {arXiv preprint arXiv:2406.09246},
  year    = {2024}
}

@article{brohan2023rt2,
  title   = {{RT-2}: Vision-Language-Action Models Transfer Web Knowledge to Robotic Control},
  author  = {Brohan, Anthony and Brown, Noah and Carbajal, Justice and Chebotar, Yevgen and Chen, Xi and Choromanski, Krzysztof and Ding, Tianli and Driess, Danny and Dubey, Avinava and Finn, Chelsea and others},
  journal = {arXiv preprint arXiv:2307.15818},
  year    = {2023}
}

@article{black2024pi_0,
  title={\(\pi_0\): A Vision-Language-Action Flow Model for General Robot Control},
  author={Black, Kevin and Brown, Noah and Driess, Danny and Esmail, Adnan and Equi, Michael and Finn, Chelsea and Fusai, Niccolo and Groom, Lachy and Hausman, Karol and Ichter, Brian and others},
  journal={arXiv preprint arXiv:2410.24164},
  year={2024}
}

@article{zhu2025unified,
  title={Unified world models: Coupling video and action diffusion for pretraining on large robotic datasets},
  author={Zhu, Chuning and Yu, Raymond and Feng, Siyuan and Burchfiel, Benjamin and Shah, Paarth and Gupta, Abhishek},
  journal={arXiv preprint arXiv:2504.02792},
  year={2025}
}

@InProceedings{Bi_2026_CVPR,
    author    = {Bi, Hongzhe and Tan, Hengkai and Xie, Shenghao and Wang, Zeyuan and Huang, Shuhe and Liu, Haitian and Zhao, Ruowen and Feng, Yao and Xiang, Chendong and Rong, Yinze and Zhao, Hongyan and Liu, Hanyu and Su, Zhizhong and Ma, Lei and Su, Hang and Zhu, Jun},
    title     = {Motus: A Unified Latent Action World Model},
    booktitle = {Proceedings of the IEEE/CVF Conference on Computer Vision and Pattern Recognition (CVPR)},
    month     = {June},
    year      = {2026},
    pages     = {35101-35113}
}

@article{kim2026cosmos,
  title={Cosmos policy: Fine-tuning video models for visuomotor control and planning},
  author={Kim, Moo Jin and Gao, Yihuai and Lin, Tsung-Yi and Lin, Yen-Chen and Ge, Yunhao and Lam, Grace and Liang, Percy and Song, Shuran and Liu, Ming-Yu and Finn, Chelsea and others},
  journal={arXiv preprint arXiv:2601.16163},
  year={2026}
}

@article{ye2026world,
  title={World action models are zero-shot policies},
  author={Ye, Seonghyeon and Ge, Yunhao and Zheng, Kaiyuan and Gao, Shenyuan and Yu, Sihyun and Kurian, George and Indupuru, Suneel and Tan, You Liang and Zhu, Chuning and Xiang, Jiannan and others},
  journal={arXiv preprint arXiv:2602.15922},
  year={2026}
}

@article{li2026causal,
  title={Causal world modeling for robot control},
  author={Li, Lin and Zhang, Qihang and Luo, Yiming and Yang, Shuai and Wang, Ruilin and Han, Fei and Yu, Mingrui and Gao, Zelin and Xue, Nan and Zhu, Xing and others},
  journal={arXiv preprint arXiv:2601.21998},
  year={2026}
}

@article{akbari2026flash,
  title={Flash-WAM: Modality-Aware Distillation for World Action Models},
  author={Akbari, Arman and Zhang, Ci and Akbari, Arash and Zhao, Lin and Chen, Yixiao and Chen, Weiwei and Zhang, Xuan and Yuan, Geng and Wang, Yanzhi},
  journal={arXiv preprint arXiv:2606.05254},
  year={2026}
}

@article{li2026efficient,
  title={Efficient-WAM: A 1B-Parameter World-Action Model with Low-Cost Future Imagination},
  author={Li, Jiajun and Guo, Tiecheng and Ye, Yifan and Zhang, Rongyu and Chi, Xiaowei and Sun, Qianpu and Li, Ying and Lou, Yunfan and Huang, Yan and Lu, Zhihe and others},
  journal={arXiv preprint arXiv:2606.10040},
  year={2026}
}

@article{cai2026aha,
  title={AHA-WAM: Asynchronous Horizon-Adaptive World-Action Modeling with Observation-Guided Context Routing},
  author={Cai, Jisong and Ling, Long and Chu, Shiwei and Liu, Zhongshan and Kang, Jiayue and Liang, Zhixuan and Xu, Wenjie and Mao, Yinan and Zhang, Weinan and Yang, Xiaokang and others},
  journal={arXiv preprint arXiv:2606.09811},
  year={2026}
}

@article{yuan2026fast,
  title={Fast-wam: Do world action models need test-time future imagination?},
  author={Yuan, Tianyuan and Dong, Zibin and Liu, Yicheng and Zhao, Hang},
  journal={arXiv preprint arXiv:2603.16666},
  year={2026}
}

@article{lin2026jepa,
  title={JEPA-WAM: Learning Vision-Language-Action Policies with Joint-Embedding World Modeling},
  author={Lin, Yihan and He, Jiawei and Bao, Shifeng and Zhao, Chen and Li, Yang and Wang, Xiaobo and Wang, Yan and Chi, Cheng and Zhang, Jing},
  journal={arXiv preprint arXiv:2608.09381},
  year={2026}
}

@article{chen2026lawam,
  title={Lawam: Latent world action models for efficient dynamics-aware robot policies},
  author={Chen, Jialei and Wang, Kai and Chen, Kang and Chen, Shuaihang and Gao, Feng and Tang, Wenhao and Li, Zhiyuan and Liu, Weilin and Yao, Zhuyu and Li, Boxun and others},
  journal={arXiv preprint arXiv:2606.15768},
  year={2026}
}

@article{wan2025wan,
  title={Wan: Open and advanced large-scale video generative models},
  author={Wan, Team and Wang, Ang and Ai, Baole and Wen, Bin and Mao, Chaojie and Xie, Chen-Wei and Chen, Di and Yu, Feiwu and Zhao, Haiming and Yang, Jianxiao and others},
  journal={arXiv preprint arXiv:2503.20314},
  year={2025}
}

@article{kim2025fine,
  title={Fine-tuning vision-language-action models: Optimizing speed and success},
  author={Kim, Moo Jin and Finn, Chelsea and Liang, Percy},
  journal={arXiv preprint arXiv:2502.19645},
  year={2025}
}

@article{pertsch2025fast,
  title={Fast: Efficient action tokenization for vision-language-action models},
  author={Pertsch, Karl and Stachowicz, Kyle and Ichter, Brian and Driess, Danny and Nair, Suraj and Vuong, Quan and Mees, Oier and Finn, Chelsea and Levine, Sergey},
  journal={arXiv preprint arXiv:2501.09747},
  year={2025}
}

@inproceedings{wang2026unified,
  title={Unified vision-language-action model},
  author={Wang, Yuqi and Li, Xinghang and Wang, Wenxuan and Zhang, Junbo and Li, Yingyan and Chen, Yuntao and Wang, Xinlong and Zhang, Zhaoxiang},
  booktitle={International Conference on Learning Representations},
  volume={2026},
  pages={80929--80944},
  year={2026}
}

@article{guo2024prediction,
  title={Prediction with action: Visual policy learning via joint denoising process},
  author={Guo, Yanjiang and Hu, Yucheng and Zhang, Jianke and Wang, Yen-Jen and Chen, Xiaoyu and Lu, Chaochao and Chen, Jianyu},
  journal={Advances in Neural Information Processing Systems},
  volume={37},
  pages={112386--112410},
  year={2024}
}

@article{team2024octo,
  title={Octo: An open-source generalist robot policy},
  author={Team, Octo Model and Ghosh, Dibya and Walke, Homer and Pertsch, Karl and Black, Kevin and Mees, Oier and Dasari, Sudeep and Hejna, Joey and Kreiman, Tobias and Xu, Charles and others},
  journal={arXiv preprint arXiv:2405.12213},
  year={2024}
}

@article{cen2025worldvla,
  title={Worldvla: Towards autoregressive action world model},
  author={Cen, Jun and Yu, Chaohui and Yuan, Hangjie and Jiang, Yuming and Huang, Siteng and Guo, Jiayan and Li, Xin and Song, Yibing and Luo, Hao and Wang, Fan and others},
  journal={arXiv preprint arXiv:2506.21539},
  year={2025}
}

@article{liu2023libero,
  title={Libero: Benchmarking knowledge transfer for lifelong robot learning},
  author={Liu, Bo and Zhu, Yifeng and Gao, Chongkai and Feng, Yihao and Liu, Qiang and Zhu, Yuke and Stone, Peter},
  journal={Advances in Neural Information Processing Systems},
  volume={36},
  pages={44776--44791},
  year={2023}
}

@article{fei2025libero,
  title={Libero-plus: In-depth robustness analysis of vision-language-action models},
  author={Fei, Senyu and Wang, Siyin and Shi, Junhao and Dai, Zihao and Cai, Jikun and Qian, Pengfang and Ji, Li and He, Xinzhe and Zhang, Shiduo and Fei, Zhaoye and others},
  journal={arXiv preprint arXiv:2510.13626},
  year={2025}
}

@article{ali2025world,
  title={World simulation with video foundation models for physical ai},
  author={Ali, Arslan and Bai, Junjie and Bala, Maciej and Balaji, Yogesh and Blakeman, Aaron and Cai, Tiffany and Cao, Jiaxin and Cao, Tianshi and Cha, Elizabeth and Chao, Yu-Wei and others},
  journal={arXiv preprint arXiv:2511.00062},
  year={2025}
}

@article{lipman2022flow,
  title={Flow matching for generative modeling},
  author={Lipman, Yaron and Chen, Ricky TQ and Ben-Hamu, Heli and Nickel, Maximilian and Le, Matt},
  journal={arXiv preprint arXiv:2210.02747},
  year={2022}
}

@article{agarwal2026cosmos,
  title={Cosmos 3: Omnimodal world models for physical ai},
  author={Agarwal, Niket and Ali, Arslan and Allen, Jon and Antolini, Martin and Aubame, Adeline and Azzolini, Alisson and Bai, Junjie and Bala, Maciej and Balaji, Yogesh and Bapst, Josh and others},
  journal={arXiv preprint arXiv:2606.02800},
  year={2026}
}

\end{document}